\documentclass[runningheads,a4paper]{llncs}
\usepackage{llncsdoc}
\usepackage{amsmath}
\usepackage{amssymb}
\usepackage[linesnumbered,ruled]{algorithm2e}
\usepackage{appendix}
\usepackage[colorlinks = true,linkcolor = blue,urlcolor =blue, citecolor = blue, anchorcolor = blue]{hyperref}
\usepackage{stmaryrd}
\usepackage{tikz}
\usetikzlibrary{bayesnet}

\usepackage{algorithmic}
\everymath{\displaystyle\everymath{}}
\usepackage{marvosym}

\begin{document}
\title{Multilevel Modeling with Structured Penalties for Classification from Imaging Genetics data}
\titlerunning{Multilevel modeling for classification from imaging genetics data} 
\author{Pascal Lu$^{1,2 (\text{\Letter})}$, Olivier Colliot$^{1,2}$, and \\ the Alzheimer's Disease Neuroimaging Initiative 
\institute{
1. Sorbonne Universit\'es, UPMC Universit\'e Paris 06, Inserm, CNRS, Institut du cerveau et la moelle (ICM), AP-HP - H\^opital Piti\'e-Salp\^etri\`ere, \\ Boulevard de l'h\^opital, 75013, Paris, France \\ \href{mailto:pascal.lu@inria.fr}{\textcolor{black}{\texttt{pascal.lu@inria.fr}}} \\
2. INRIA Paris, ARAMIS project-team, 75013, Paris, France \\
\href{mailto:olivier.colliot@upmc.fr}{\textcolor{black}{\texttt{olivier.colliot@upmc.fr}}}}
}
\authorrunning{P. Lu et {al.}}

\maketitle

\begin{abstract}
In this paper, we propose a framework for automatic classification of patients from multimodal genetic and brain imaging data by optimally combining them. Additive models with unadapted penalties (such as the classical group lasso penalty or $\ell_1$-multiple kernel learning) treat all modalities in the same manner and can result in undesirable elimination of specific modalities when their contributions are unbalanced. 
To overcome this limitation, we introduce a multilevel model that combines imaging and genetics and that considers joint effects between these two modalities for diagnosis prediction. Furthermore, we propose a framework allowing to combine several penalties taking into account the structure of the different types of data, such as a group lasso penalty over the genetic modality and a $\ell_2$-penalty on imaging modalities. Finally, we propose a fast optimization algorithm, based on a proximal gradient method. 
The model has been evaluated on genetic (single nucleotide polymorphisms - SNP) and imaging (anatomical MRI measures) data from the ADNI database, and compared to additive models \cite{wang_identifying_2012,aiolli_easymkl_2014}. It exhibits good performances in AD diagnosis; and at the same time, reveals relationships between genes, brain regions and the disease status.
\end{abstract}


\section{Introduction}

The research area of imaging genetics studies the association between genetic and brain imaging data \cite{liu_review_2014}. A large number of papers studied the relationship between genetic and neuroimaging data by considering that a phenotype can be explained by a sum of effects from genetic variants. These multivariate approaches use partial least squares \cite{lorenzi_2016}, sparse canonical correlation analysis \cite{du_novel_2014}, sparse regularized linear regression with a $\ell_1$-penalty \cite{kohannim_discovery_2012}, group lasso penalty \cite{silver_identification_2012,silver_fast_2012}, or Bayesian model that links genetic variants to imaging regions and imaging regions to the disease status \cite{batmanghelich_probabilistic_2016}.

But another interesting problem is about combining genetic and neuroimaging data for automatic classification of patients. In particular, machine learning methods have been used to build predictors for heterogeneous data, coming from different modalities for brain disease diagnosis, such as Alzheimer's disease (AD) diagnosis. However, challenging issues are high-dimensional data, small number of observations, the heterogeneous nature of data, and the weight for each modality. 

A framework that is commonly used to combine heterogeneous data is multiple kernel learning (MKL) \cite{gonen_multiple_2011}. In MKL, each modality is represented by a kernel (usually a linear kernel). The decision function and weights for the kernel are simultaneously learnt. Moreover, the group lasso \cite{ming_model_2006,meier_group_2008} is a way to integrate structure inside data. However, the standard $\ell_1$-MKL and group lasso may eliminate modalities that have a weak contribution. In particular, for AD, imaging data already provides good results for its diagnosis. To overcome this problem, different papers have proposed to use a $\ell_{1,p}$-penalty \cite{kloft_lp-norm_2011} to combine optimally different modalities \cite{wang_identifying_2012,peng_structured_2016}. 

These approaches do not consider potential effects between genetic and imaging data for diagnosis prediction, as they only capture brain regions and SNPs separately taken. Moreover, they put on the same level genetic and imaging data, although these data do not provide the same type of information: given only APOE genotyping, subjects can be classified according to their risk to develop AD in the future; on the contrary, imaging data provides a photography of the subject's state at the present time. 

Thereby, we propose a new framework that makes hierarchical the parameters and considers interactions between genetic and imaging data for AD diagnosis. We started with the idea that learning AD diagnosis from imaging data already provides good results. Then, we considered that the decision function parameters learnt from imaging data could be modulated, depending on each subject's genetic data. In other words, genes would express themselves through these parameters. Considering a linear regression that links these parameters and the genetic data, it leads to a multilevel model between imaging and genetics. Our method also proposes potential relations between genetic and imaging variables, if both of them are simultaneously related to AD. This approach is different from the modeling proposed by \cite{batmanghelich_probabilistic_2016}, where imaging variables are predicted from genetic variables, and diagnosis is predicted from imaging variables.

Furthermore, current approaches \cite{wang_identifying_2012,peng_structured_2016,aiolli_easymkl_2014} do not exploit data structure inside each modality, as it is logical to group SNPs by genes, to expect sparsity between genes (all genes are not linked to AD) and to enforce a smooth regularization over brain regions for imaging modality. Thus, we have imposed specific penalties for each modality by using a $\ell_2$-penalty on the imaging modality, and a group lasso penalty over the genetic modality. It models the mapping of variants into genes, providing a better understanding of the role of genes in AD.

To learn all the decision function parameters, a fast optimization algorithm, based on a proximal gradient method, has been developed. Finally, we have evaluated our model on 1,107 genetic (SNP) and 114 imaging (anatomical MRI measures) variables from the ADNI database\footnote{\href{http://adni.loni.usc.edu}{http://adni.loni.usc.edu}} and compared it to additive models \cite{wang_identifying_2012,aiolli_easymkl_2014}.

\section{Model set-up}

\subsection{Multilevel Logistic Regression with Structured Penalties}\label{multdescr}

Let $\{ (\mathbf{x}_{\mathcal{G}}^k, \mathbf{x}_{\mathcal{I}}^k, y^k), k = 1, \ldots, N\}$ be a set of labeled data, with $\mathbf{x}_{\mathcal{G}}^k \in \mathbb{R}^{|\mathcal{G}|}$ (genetic data), and $\mathbf{x}_{\mathcal{I}}^k \in \mathbb{R}^{|\mathcal{I}|}$ (imaging data) and $y^k \in \{0, 1\}$ (diagnosis). Genetic, imaging and genetic-imaging cross products training data are assumed centered and normalized.

We propose the following Multilevel Logistic Regression model:
\[ p(y=1 | \mathbf{x}_{\mathcal{G}}, \mathbf{x}_{\mathcal{I}}) =\sigma\left( \boldsymbol{\alpha}(\mathbf{x}_{\mathcal{G}})^{\top} \mathbf{x}_{\mathcal{I}}+\alpha_0(\mathbf{x}_{\mathcal{G}})\right) \quad\text{with}~\sigma: x \mapsto \frac{1}{1+\text{e}^{-x}} \]

where $\alpha_0(\mathbf{x}_{\mathcal{G}})$ is the intercept and $\boldsymbol{\alpha}(\mathbf{x}_{\mathcal{G}}) \in \mathbb{R}^{|\mathcal{I}|}$ is the parameter vector. On the contrary of the classical logistic regression model, we propose a multilevel model, for which the parameter vector $\boldsymbol{\alpha}(\mathbf{x}_{\mathcal{G}})$ and the intercept $\alpha_0(\mathbf{x}_{\mathcal{G}})$ depend on genetic data $\mathbf{x}_{\mathcal{G}}$. 

\begin{figure}[ht!]
\begin{center}
\scalebox{1}{
\begin{tikzpicture}
  \node[obs]                               (y) {$y$};
  \node[latent, above=of y, xshift=0cm,yshift=-0.5cm] (alpha) {$\boldsymbol{\alpha}(\mathbf{x}_{\mathcal{G}})$};
  \node[obs, above=of y, xshift=3cm,yshift=-0.5cm]  (xI) {$\mathbf{x}_{\mathcal{I}}$};
  \node[latent, above=of y, xshift=-3cm,yshift=-0.5cm] (alpha0) {$\alpha_0(\mathbf{x}_{\mathcal{G}})$};
  
  \node[latent, above=of alpha, xshift=-4.5cm,yshift=-0.5cm] (beta0G) {$\beta_0$};
  \node[latent, above=of alpha, xshift=-3cm,yshift=-0.5cm] (betaG) {$\boldsymbol{\beta}_{\mathcal{G}}$};
  \node[obs, above=of alpha,xshift=-1.5cm,yshift=-0.5cm]  (xG) {$\mathbf{x}_{\mathcal{G}}$};
  \node[latent, above=of alpha, xshift=0cm,yshift=-0.5cm] (betaI) {$\boldsymbol{\beta}_{\mathcal{I}}$};
  \node[latent, above=of alpha, xshift=1.5cm,yshift=-0.5cm] (W) {$\mathbf{W}$};
  \edge {xI,alpha} {y} ; %
  \edge {xG} {alpha,alpha0} ; %
  \edge {betaI,W} {alpha} ; %
  \edge {beta0G,betaG} {alpha0} ; %
  \edge {alpha0} {y} ; %
\end{tikzpicture}
}
\end{center}
\caption{The disease status $y$ is predicted from imaging data $\mathbf{x}_{\mathcal{I}}$ and the parameters $\beta_0(\mathbf{x}_{\mathcal{G}}), \boldsymbol{\beta}(\mathbf{x}_{\mathcal{G}})$ (which are computed from genetic data $\mathbf{x}_{\mathcal{G}}$)}
\label{relationship}
\end{figure}
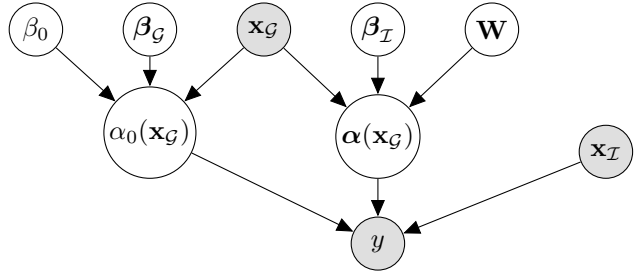

This is to be compared to an additive model, where the diagnosis is directly deduced from genetic and imaging data put at the same level. We assume that $\boldsymbol{\alpha}$ and $\alpha_0$ are affine functions of genetic data $\mathbf{x}_{\mathcal{G}}$:
\[\boldsymbol{\alpha}(\mathbf{x}_{\mathcal{G}}) = \mathbf{W}\mathbf{x}_{\mathcal{G}} + \boldsymbol{\beta}_{\mathcal{I}} \quad \text{and} \quad \alpha_0(\mathbf{x}_{\mathcal{G}}) = \boldsymbol{\beta}_{\mathcal{G}}^{\top}\mathbf{x}_{\mathcal{G}} + \beta_0 \]

where $\mathbf{W} \in \mathcal{M}_{|\mathcal{I}|, |\mathcal{G}|}(\mathbb{R})$, $\boldsymbol{\beta}_{\mathcal{I}} \in \mathbb{R}^{|\mathcal{I}|}$, $\boldsymbol{\beta}_{\mathcal{G}} \in \mathbb{R}^{|\mathcal{G}|}$ and $\beta_0 \in \mathbb{R}$. Therefore, the probability becomes $p( y=1 | \mathbf{x}_{\mathcal{G}}, \mathbf{x}_{\mathcal{I}}) = \sigma\left(\mathbf{x}_{\mathcal{G}}^{\top}\mathbf{W}^{\top}\mathbf{x}_{\mathcal{I}} + \boldsymbol{\beta}_{\mathcal{I}}^{\top}\mathbf{x}_{\mathcal{I}} + \boldsymbol{\beta}_{\mathcal{G}}^{\top}\mathbf{x}_{\mathcal{G}} + \beta_0\right)$. Figure \ref{relationship} summarizes the relations between parameters. 
\medskip

\medskip

The parameters $\mathbf{W}, \boldsymbol{\beta}_{\mathcal{I}}, \boldsymbol{\beta}_{\mathcal{G}}, \beta_0$ are obtained by minimizing the objective:
\begin{align*}
S(\mathbf{W}, \boldsymbol{\beta}_{\mathcal{I}}, \boldsymbol{\beta}_{\mathcal{G}}, \beta_0) & = R_N(\mathbf{W}, \boldsymbol{\beta}_{\mathcal{I}}, \boldsymbol{\beta}_{\mathcal{G}}, \beta_0) + \Omega(\mathbf{W}, \boldsymbol{\beta}_{\mathcal{I}}, \boldsymbol{\beta}_{\mathcal{G}}) \\
\label{risk}
\text{with } R_N(\mathbf{W}, \boldsymbol{\beta}_{\mathcal{I}}, \boldsymbol{\beta}_{\mathcal{G}}, \beta_0) & = \frac{1}{N}\sum_{k=1}^N \left\{ -  y^k\left( {(\mathbf{x}_{\mathcal{G}}^{k})}^{\top} \mathbf{W}^{\top}\mathbf{x}_{\mathcal{I}}^k + \boldsymbol{\beta}_{\mathcal{I}}^{\top}\mathbf{x}_{\mathcal{I}}^k + \boldsymbol{\beta}_{\mathcal{G}}^{\top}\mathbf{x}_{\mathcal{G}}^k + \beta_0\right) \right. \\
&\quad\quad\quad\quad\quad\quad \left. + \log \left(1+\text{e}^{(\mathbf{x}_{\mathcal{G}}^k)^{\top}\mathbf{W}^{\top}\mathbf{x}_{\mathcal{I}}^k + \boldsymbol{\beta}_{\mathcal{I}}^{\top}\mathbf{x}_{\mathcal{I}}^k + \boldsymbol{\beta}_{\mathcal{G}}^{\top}\mathbf{x}_{\mathcal{G}}^k + \beta_0}\right) \right\} \\
\text{and }\quad\quad \Omega(\mathbf{W}, \boldsymbol{\beta}_{\mathcal{I}}, \boldsymbol{\beta}_{\mathcal{G}}) & =  \lambda_{W}  \Omega_W ( \mathbf{W} )+ \lambda_{\mathcal{I}}  \Omega_{\mathcal{I}} ( \boldsymbol{\beta}_{\mathcal{I}} ) + \lambda_{\mathcal{G}} \Omega_{\mathcal{G}} ( \boldsymbol{\beta}_{\mathcal{G}} )\end{align*}

$\Omega_W$, $\Omega_{\mathcal{I}}$, $\Omega_{\mathcal{G}}$ are respectively the penalties for $\mathbf{W}$, $\boldsymbol{\beta}_{\mathcal{I}}$, $\boldsymbol{\beta}_{\mathcal{G}}$, whereas $\lambda_{W} >0$, $\lambda_{\mathcal{I}} > 0$, $\lambda_{\mathcal{G}} >0$ are respectively the regularization parameters for $\Omega_W$, $\Omega_{\mathcal{I}}$, $\Omega_{\mathcal{G}}$. 

Genetic data are a sequence of single-polymorphism nucleotides (SNP) counted by minor allele. A SNP can belong (or not) to one gene ${\ell}$ (or more) and therefore participate in the production of proteins that interact inside pathways. We decided to group SNPs by genes, and designed a penalty  to enforce sparsity between genes and regularity inside genes. Given that some SNPs may belong to multiple genes, the group lasso with overlap penalty \cite{jacob_group_2009} is more suitable, with genes as groups. To deal with this penalty, an overlap expansion is performed. Given $\mathbf{x} \in \mathbb{R}^{|\mathcal{G}|}$ a subject's feature vector, a new feature vector is created $\widetilde{\mathbf{x}} = \left( \mathbf{x}_{\mathcal{G}_1}^{\top}, \ldots, \mathbf{x}_{\mathcal{G}_L}^{\top} \right)^{\top} \in \mathbb{R}^{\sum_{{\ell}=1}^L |\mathcal{G}_{\ell}|}$, defined by the concatenation of copies of the genetic data restricted by group $\mathcal{G}_{\ell}$. Similarly, the same expansion is performed on ${\boldsymbol{\beta}}_{\mathcal{G}}, \mathbf{W}$ to obtain $\widetilde{\boldsymbol{\beta}}_{\mathcal{G}} \in \mathbb{R}^{\sum_{{\ell}=1}^L |\mathcal{G}_{\ell}|}$ and $\widetilde{\mathbf{W}} \in \mathbb{R}^{|\mathcal{I}|\times(\sum_{{\ell}=1}^L |\mathcal{G}_{\ell}|)}$. This group lasso with overlap penalty is used for the matrix $\mathbf{W}$ and for ${\boldsymbol{\beta}}_{\mathcal{G}}$.

For imaging variables, the ridge penalty is considered: $\Omega_{\mathcal{I}}(\boldsymbol{\beta}_{\mathcal{I}}) = \left\|\boldsymbol{\beta}_{\mathcal{I}}\right\|_2^2$. In particular, brain diseases usually have a diffuse anatomical pattern of alteration throughout the brain and therefore, regularity is usually required for the imaging parameter. Finally, $\Omega$ is defined by:
\[  \Omega \left(\widetilde{\mathbf{W}}, \widetilde{\boldsymbol{\beta}}_{\mathcal{G}}, \boldsymbol{\beta}_{\mathcal{I}}\right) = \lambda_{W} \sum_{i=1}^{|\mathcal{I}|} \sum_{\ell=1}^{L}\theta_{\mathcal{G}_{\ell}} \left\|\widetilde{\mathbf{W}}_{i, \mathcal{G}_{\ell}}\right\|_2 +\lambda_{\mathcal{I}} \left\| \widetilde{\boldsymbol{\beta}}_{\mathcal{I}}\right\|_2 + \lambda_{\mathcal{G}} \sum_{{\ell}=1}^L \theta_{\mathcal{G}_{\ell}} \left\| \widetilde{\boldsymbol{\beta}}_{\mathcal{G}_{\ell}} \right\|_2   \]

\subsection{Minimization of $S(\mathbf{W}, \boldsymbol{\beta}_{\mathcal{I}}, \boldsymbol{\beta}_{\mathcal{I}}, \beta_0)$}

From now on, and for simplicity reasons, $\widetilde{\mathbf{W}}$, $\widetilde{\boldsymbol{\beta}}$ and $\widetilde{\mathbf{x}}$ are respectively denoted as $\mathbf{W}$, $\boldsymbol{\beta}$ and ${\mathbf{x}}$. Let $\Phi$ be the function that reshapes a matrix of $\mathcal{M}_{|\mathcal{I}|, |\mathcal{G}|}(\mathbb{R})$ to a vector of $\mathbb{R}^{|\mathcal{I}| \times |\mathcal{G}|}$ (i.e. $\mathbf{W}_{i, g} = \Phi(\mathbf{W})_{i |\mathcal{G}| + g}$):
\[ \Phi: \mathbf{W} \mapsto ((\mathbf{W}_{1, 1}, \ldots, \mathbf{W}_{1, |\mathcal{G}|}), \ldots, (\mathbf{W}_{|\mathcal{I}|, 1}, \ldots, \mathbf{W}_{|\mathcal{I}|, |\mathcal{G}|}))\]

We will estimate $\Phi(\mathbf{W})$ and then reshape it to obtain $\mathbf{W}$. The algorithm developed is based on a proximal gradient method \cite{hastie_statistical_2015,beck_gradient_2010}. 

The parameters $\mathbf{w}^{(t+1)} = \left( \Phi\left( \mathbf{W}^{(t+1)}\right), \boldsymbol{\beta}_{\mathcal{I}}^{(t+1)}, \boldsymbol{\beta}_{\mathcal{G}}^{(t+1)}, {\beta}_0^{(t+1)}\right)$ are updated with:
\begin{align*} \mathbf{w}^{(t+1)} &= \mathop{\text{argmin}}_{\mathbf{w} } R_N\left( \mathbf{w} \right) + \left[\mathbf{w} - \mathbf{w}^{(t)}  \right]^{\top} \nabla R_N \left(\mathbf{w}^{(t)}\right) +\frac{1}{2\varepsilon} \left\|\mathbf{w} - \mathbf{w}^{(t)}\right\|_2^2 + \Omega(\mathbf{w}) \\
 &= \mathop{\text{argmin}}_{\mathbf{w} } \left\{ \frac{1}{2} \left\| 
\boldsymbol{\omega}^{(t)} -\mathbf{w}^{(t)}\right\|_2^2 + \varepsilon\Omega(\mathbf{w}) \right\}
\text{ with } \boldsymbol{\omega}^{(t)} = \mathbf{w}^{(t)} - \varepsilon \nabla R_N\left(\mathbf{w}^{(t)}\right) \end{align*}

The idea is to update $\mathbf{w}^{(t+1)}$ from $\mathbf{w}^{(t)}$ with a Newton-type algorithm without the constraint $\Omega$ given a stepsize $\varepsilon$, and then to project the result onto the compact set defined by $\Omega$. Regarding the stepsize $\varepsilon$, a backtracking line search \cite{beck_gradient_2010} is performed. Let $\widehat{G}\left(\mathbf{w}^{(t)}, \varepsilon\right)  = \frac{1}{\varepsilon} \left[  \mathbf{w}^{(t)} - \mathbf{w}^{(t+1)} \right]$ be the step in the proximal gradient update. A line search is performed over $\varepsilon$ until the inequality is reached:
\begin{align*}
R_N\left(\mathbf{w}^{(t+1)}\right) & \leq R_N\left( \mathbf{w}^{(t)} \right)  - \varepsilon\nabla R_N\left(\mathbf{w}^{(t)}\right)^{\top} \widehat{G}\left(\mathbf{w}^{(t)}, \varepsilon\right) +\frac{\varepsilon}{2} \left\|\widehat{G}\left(\mathbf{w}^{(t)}, \varepsilon\right)\right\|_2^2 \end{align*}

The minimization algorithm stops when {\small{$\left| S\left(\mathbf{w}^{(t+1)} \right) - S\left(\mathbf{w}^{(t)}\right) \right| \leq \eta \left| S\left(\mathbf{w}^{(t)}\right) \right| $}}, where $\eta = 10^{-5}$.  The whole algorithm is summarized below: 

\begin{algorithm}[ht!]
\caption{Training the multilevel logistic regression}\label{proximal-update}

\textbf{Input}: $\{ (\mathbf{x}_{\mathcal{I}}^k,  \mathbf{x}_{\mathcal{G}}^k,    y^k),   k = 1,   \ldots,   N \}$,   $\delta= 0.8$,   $\varepsilon_0 = 1$,   $\eta = 10^{-5}$ \;
\textbf{Initialization}: $\mathbf{W} = \mathbf{0}$,   $\boldsymbol{\beta}_{\mathcal{I}} = \mathbf{0}$,    $\boldsymbol{\beta}_{\mathcal{G}} = \mathbf{0}$,   $\beta_0=0$ and continue = True \;
 \While{\textnormal{continue}}{ 
$\varepsilon = \varepsilon_0$\;

$R_N = R_N\left({\mathbf{W}},   {\boldsymbol{\beta}}_{\mathcal{I}},   {\boldsymbol{\beta}}_{\mathcal{G}},   {\beta}_0 \right)$\;

$\displaystyle{\nabla R_N = \frac{1}{N}\sum_{k=1}^N \begin{pmatrix} \Phi\left({(\mathbf{x}_{\mathcal{I}}^{k})}^{\top} \mathbf{x}_{\mathcal{G}}^k\right) \\ \mathbf{x}_{\mathcal{I}}^k \\ \mathbf{x}_{\mathcal{G}}^k \\ 1\end{pmatrix} \left[ \sigma \left({(\mathbf{x}_{\mathcal{G}}^{k})}^{\top} \mathbf{W}^{\top}\mathbf{x}_{\mathcal{I}}^k + \boldsymbol{\beta}_{\mathcal{I}}^{\top}\mathbf{x}_{\mathcal{I}}^k + \boldsymbol{\beta}_{\mathcal{G}}^{\top}\mathbf{x}_{\mathcal{G}}^k + \beta_0\right) -y^k  \right]}$

$\left( \widehat{\mathbf{W}},   \widehat{\boldsymbol{\beta}}_{\mathcal{I}},   \widehat{\boldsymbol{\beta}}_{\mathcal{G}},   \widehat{\beta}_0,   \widehat{G}\right) = $ \textbf{Algo\_\ref{proximal-parameter-update}}$(
{\mathbf{W}},   {\boldsymbol{\beta}}_{\mathcal{I}},   {\boldsymbol{\beta}}_{\mathcal{G}},   {\beta}_0,   \nabla R_N,    \varepsilon)$\;

\While{$R_N\left(\widehat{\mathbf{W}},   \widehat{\boldsymbol{\beta}}_{\mathcal{I}},   \widehat{\boldsymbol{\beta}}_{\mathcal{G}},   \widehat{\beta}_0 \right) > R_N  - \varepsilon\nabla R_N^{\top} \widehat{G} +\frac{\varepsilon}{2} \|\widehat{G}\|_2^2$}{
$\varepsilon = \delta \varepsilon$ and $\left( \widehat{\mathbf{W}},   \widehat{\boldsymbol{\beta}}_{\mathcal{I}},   \widehat{\boldsymbol{\beta}}_{\mathcal{G}},   \widehat{\beta}_0,   \widehat{G}\right) = $ \textbf{Algo\_\ref{proximal-parameter-update}}$(
{\mathbf{W}},   {\boldsymbol{\beta}}_{\mathcal{I}},   {\boldsymbol{\beta}}_{\mathcal{G}},   {\beta}_0,   \nabla R_N,    \varepsilon)$\;
}
continue = $\left| S\left(\widehat{\mathbf{W}},   \widehat{\boldsymbol{\beta}}_{\mathcal{I}},   \widehat{\boldsymbol{\beta}}_{\mathcal{G}},   \widehat{\beta}_0\right) - S\left(\mathbf{W},   {\boldsymbol{\beta}}_{\mathcal{I}},   {\boldsymbol{\beta}}_{\mathcal{G}},   \beta_0\right) \right| \mathop{>}^{\text{?}} \eta \left| S\left(\mathbf{W},   {\boldsymbol{\beta}}_{\mathcal{I}},   {\boldsymbol{\beta}}_{\mathcal{G}},   \beta_0\right) \right|$

${\mathbf{W}} = \widehat{\mathbf{W}}$,   ${\boldsymbol{\beta}}_{\mathcal{I}} = \widehat{\boldsymbol{\beta}}_{\mathcal{I}}$,   ${\boldsymbol{\beta}}_{\mathcal{G}} = \widehat{\boldsymbol{\beta}}_{\mathcal{G}}$,   $\beta_0 = \widehat{\beta}_0$\;
}
\textbf{return}  $\left(\mathbf{W},   {\boldsymbol{\beta}}_{\mathcal{I}},   {\boldsymbol{\beta}}_{\mathcal{G}},   \beta_0\right)$
\end{algorithm}

\begin{algorithm}[ht!]
\caption{Parameter update}\label{proximal-parameter-update}
\textbf{Input}: $({\mathbf{W}},   {\boldsymbol{\beta}}_{\mathcal{I}},   {\boldsymbol{\beta}}_{\mathcal{G}},   {\beta}_0)$ (parameters),   $\nabla R_N$ (gradient),   $\varepsilon$ (stepsize) \;

Compute $\boldsymbol{\omega} = \boldsymbol{\beta}  - \varepsilon \nabla_{({\mathbf{W}},   {\boldsymbol{\beta}}_{\mathcal{I}},   {\boldsymbol{\beta}}_{\mathcal{G}})} R_N$ \;

Update $\widehat{\mathbf{W}}_{\mathcal{G}_{\ell},   i} = \max\left( 0,   1 - \frac{\varepsilon \lambda_{\mathcal{G}} \theta_{\mathcal{G}_{\ell}}}{\left\| \boldsymbol{\omega}^{(t)}_{\mathcal{G}_{\ell} + i |\mathcal{G}|}  \right\|_2} \right) \boldsymbol{\omega}_{\mathcal{G}_{\ell} + i |\mathcal{G}|}$
for $(i,   \ell) \in \llbracket 1,    |\mathcal{I}| \rrbracket \times \llbracket 1,   L \rrbracket$ \;

Update $\widehat{\boldsymbol{\beta}}_{\mathcal{I}} =  \frac{\boldsymbol{\omega}_{\mathcal{I}+ |\mathcal{G}| |\mathcal{I}|}}{1 +  2 \varepsilon \lambda_{\mathcal{I}} }$ (imaging modality) \;

Update $\widehat{\boldsymbol{\beta}}_{\mathcal{G}_{\ell}} =  \max\left(0,   1 - \frac{\varepsilon \lambda_{\mathcal{G}} \theta_{\mathcal{G}_{\ell}}}{\left\| \boldsymbol{\omega}^{(t)}_{\mathcal{G}_{\ell}+(|\mathcal{G}|+1) |\mathcal{I}|}  \right\|_2} \right) \boldsymbol{\omega}_{\mathcal{G}_{\ell}+(|\mathcal{G}|+1) |\mathcal{I}|}$ for $\ell \in \llbracket 1,   L \rrbracket$ \;

Update $\widehat{\beta}_{0} = \beta_{0} - \varepsilon\frac{\partial R_N}{\partial \beta_0}$ and $\widehat{G} = \frac{1}{\varepsilon}\left[ \begin{pmatrix} \Phi({\mathbf{W}}) \\ {\boldsymbol{\beta}}_{\mathcal{I}}  \\ {\boldsymbol{\beta}}_{\mathcal{G}}   \\ {{\beta}}_{0}  \end{pmatrix} -\begin{pmatrix} \Phi(\widehat{\mathbf{W}}) \\ \widehat{\boldsymbol{\beta}}_{\mathcal{I}}  \\ \widehat{\boldsymbol{\beta}}_{\mathcal{G}}   \\ \widehat{{\beta}}_{0}  \end{pmatrix} \right]$ \;
\textbf{return} $\left(\widehat{\mathbf{W}},   \widehat{\boldsymbol{\beta}}_{\mathcal{I}},   \widehat{\boldsymbol{\beta}}_{\mathcal{G}},   \widehat{\beta}_0,   \widehat{G} \right)$
\end{algorithm}

\section{Experimental results}
\subsection{Dataset}

The ADNI1 GWAS dataset from ADNI studied 707 subjects, with 156 Alzheimer's Disease patients (denoted AD), {196} MCI patients at baseline who progressed to AD (denoted pMCI, as progressive MCI), {150} MCI patients who remain stable (denoted sMCI, as stable MCI) and 201 healthy control subjects (denoted CN).

In ADNI1 GWAS dataset, 620,901 SNPs have been genotyped, but we selected 1,107 SNPs based on the 44 first top genes related to AD (from AlzGene\footnote{\href{http://www.alzgene.org}{http://www.alzgene.org}}) and on the Illumina annotation using the Genome build 36.2. Group weighting for genes is based on gene size: for group $\mathcal{G}_{\ell}$, the weight $\theta_{\mathcal{G}_{\ell}}=\sqrt{|\mathcal{G}_{\ell}|}$ ensures that the penalty term is of the order of the number of parameters of the group. 

The parameter $\lambda_{\mathcal{G}}$ influences the number of groups that are selected by the model. In particular, the group $\mathcal{G}_{\ell}$ enters in the model during the first iteration if $\left\| \nabla_{\boldsymbol{\beta}_{\mathcal{G}_{\ell}}} R_N(\mathbf{0}) \right\|_2 > \lambda_{\mathcal{G}} \theta_{\mathcal{G}_{\ell}}$. This inequality gives an upper bound for $\lambda_{\mathcal{G}}$. The same remark can be done for $\lambda_W$. 
Regarding MRI modality, we used the segmentation of FreeSurfer which gives the volume of subcortical regions (44 features) and the average cortical region thickness (70 features). Therefore, there are $1,107\times 114 = 126,198$ parameters to infer for $\mathbf{W}$, $114$ parameters for $\boldsymbol{\beta}_{\mathcal{I}}$ and $1,107$ parameters for $\boldsymbol{\beta}_{\mathcal{G}}$.

\subsection{Results}

We ran our multilevel model and compared it to the logistic regression applied to one single modality with simple penalties (lasso, group lasso, ridge), to additive models ({\cite{wang_identifying_2012}}, \cite{aiolli_easymkl_2014} EasyMKL with a linear kernel for each modality, and the model $p(y=1 | \mathbf{x}_{\mathcal{G}}, \mathbf{x}_{\mathcal{I}}) = \sigma\left(  \boldsymbol{\beta}_{\mathcal{I}}^{\top}\mathbf{x}_{\mathcal{I}} + \boldsymbol{\beta}_{\mathcal{G}}^{\top}\mathbf{x}_{\mathcal{G}} + \beta_0\right)$ with our algorithm under the constraint $\boldsymbol{\beta}_{\mathcal{G}} \neq \mathbf{0}$), and to the multiplicative model with $\mathbf{W}$ only, where $p( y=1 | \mathbf{x}_{\mathcal{G}}, \mathbf{x}_{\mathcal{I}}) = \sigma\left( \mathbf{x}_{\mathcal{G}}^{\top}\mathbf{W}^{\top}\mathbf{x}_{\mathcal{I}} + \beta_0\right)$. We considered two classification tasks: ``AD versus CN" and ``pMCI versus CN". Four measures are used: the sensitivity (\textsc{Sen}), the specificity (\textsc{Spe}), the precision (\textsc{Pre}) and the balanced accuracy between the sensitivity and the specificity (\textsc{BAcc}). A 10-fold cross validation is performed. The parameters $\lambda_{W}, \lambda_{\mathcal{I}}, \lambda_{\mathcal{G}}$ are optimised between $[10^{-3}, 1]$. Classification results for these tasks are shown on table \ref{table1}. It typically takes between 5 and 8 minutes to learn the parameters.

\begin{table}[ht!]
\caption{Classification results for different modalities and methods}\label{table1}
\begin{center}
\begin{tabular}{|c|c|c|c|c|c|c|} \cline{3-6}
\multicolumn{2}{c|}{ }  & \multicolumn{4}{c|}{\textsc{AD versus CN} (\%)} \\ \hline
\textsc{Modality} & \textsc{Method} \& \textsc{Penalty} & ~\textsc{Sen}~ & ~\textsc{Spe}~ & ~\textsc{Pre}~ & \textsc{BAcc} \\ 
\hline
SNPs only & logistic regression (lasso $\ell_1$) & 69.4  & 77.5 & 71.1  & 73.4 \\  
SNPs grouped by genes& logistic regression (group lasso) &  69.4  & 77.5 & 71.1  & 73.4 \\     
MRI (cortical) & logistic regression (ridge $\ell_2$) & 84.4  & 89.5 & 87.1 &  86.9 \\ 
MRI (subcortical) & logistic regression (ridge $\ell_2$) & 80.0   & 86.0  & 83.2 &  83.0 \\ 
\hline
SNP + MRI (all) & {\cite{aiolli_easymkl_2014}} EasyMKL, \emph{Aiolli et al.} & 89.4  & 85.0  & 83.0  & 87.2  \\  
{SNP + MRI (all)} &{\cite{wang_identifying_2012}} \emph{Wang et al.} & 89.4   & 88.0   & 85.7  & 88.7 \\  
\hline
SNP + MRI (all) & additive model ($\boldsymbol{\beta}_{\mathcal{I}}, \boldsymbol{\beta}_{\mathcal{G}}$ only) & 88.8  &  89.5 & 87.6  & 89.1 \\  
SNP + MRI (all) & multiplicative model ($\mathbf{W}$ only) & 89.4  & 87.0  & 85.0  & 88.2  \\ 
SNP + MRI (all) & multilevel model (all) & 90.6  &  87.0 &  85.5 &  88.8 \\ \hline
\end{tabular}

\medskip

\begin{tabular}{|c|c|c|c|c|c|c|} \cline{3-6}
\multicolumn{2}{c|}{ }  & \multicolumn{4}{c|}{\textsc{pMCI versus CN} (\%)} \\ \hline
\textsc{Modality} & \textsc{Method} \& \textsc{Penalty} & ~\textsc{Sen}~ & ~\textsc{Spe}~ & ~\textsc{Pre}~ & \textsc{BAcc} \\ 
\hline
SNPs only & logistic regression (lasso $\ell_1$) & 72.0  & 77.0 &  75.9 & 74.5 \\  
SNPs grouped by genes& logistic regression (group lasso) & 72.0  &  77.0 &  75.9 & 74.5 \\  
MRI (cortical) & logistic regression (ridge $\ell_2$) &  74.0 & 76.0 & 76.4  & 75.0 \\  
MRI (subcortical) & logistic regression (ridge $\ell_2$)  & 73.0  &  76.5 &  76.6 & 74.7 \\ 
\hline
SNP + MRI (all) & {\cite{aiolli_easymkl_2014}} EasyMKL, \emph{Aiolli et al.} &77.0   & 73.5 &  75.1 & 75.3 \\  
{SNP + MRI (all)} &{\cite{wang_identifying_2012}} \emph{Wang et al.} &  79.5 &  81.5 &  82.4 &  80.5 \\  
\hline
SNP + MRI (all) & additive model ($\boldsymbol{\beta}_{\mathcal{I}}, \boldsymbol{\beta}_{\mathcal{G}}$ only) & 80.5   & 81.0  &  82.0 & 80.8 \\  
SNP + MRI (all) & multiplicative model ($\mathbf{W}$ only) & 81.0  &  81.5 &  82.9 & 81.3  \\ 
SNP + MRI (all) & multilevel model (all) &  82.5 &  83.0 & 84.1  & 82.8 \\ \hline
\end{tabular}
\end{center}
\end{table}

Regarding MRI features, the most important features (in weight) are the left/right hippocampus, the left/right Amygdala, the left/right entorhinal and the left middle temporal cortices. 
Regarding genetic features, the most important features in weight are SNPs that belong to gene APOE (rs429358) for both tasks ``AD versus CN" and ``pMCI versus CN".

\begin{figure}[ht!]
\begin{center}
\includegraphics[width=12.3cm]{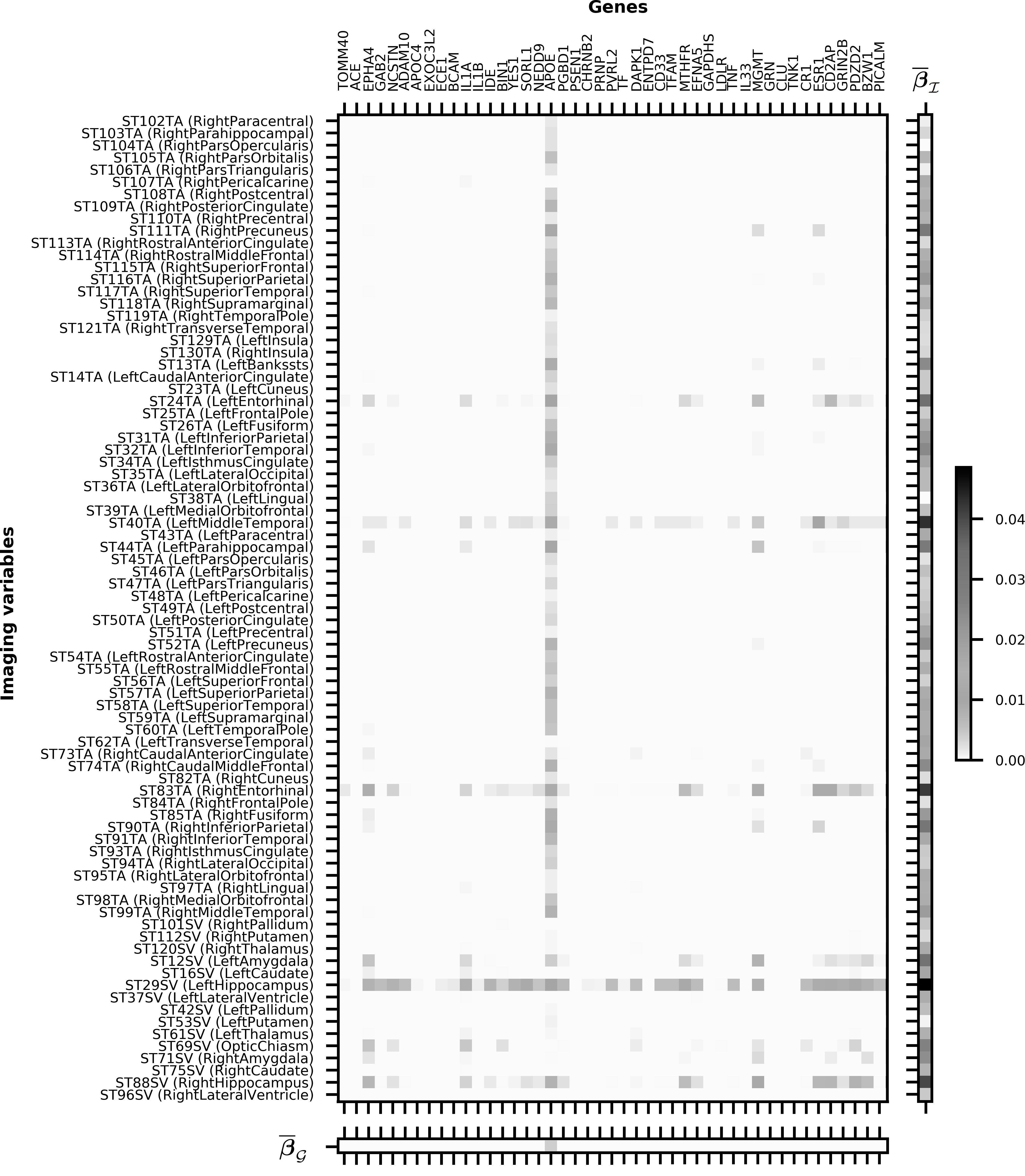}
\caption{Overview of the reduced parameters $\overline{\mathbf{W}} \in \mathbb{R}^{|\mathcal{I}| \times L}$, $\overline{\boldsymbol{\beta}}_{\mathcal{I}} \in \mathbb{R}^{|\mathcal{I}|}$ and $\overline{\boldsymbol{\beta}}_{\mathcal{G}} \in \mathbb{R}^{L}$ (learnt through the task ``pMCI vs CN" for the whole model). For brain region $i$ and gene ${\ell}$, $\overline{\mathbf{W}}[i, {\ell}] = \max_{g \in \mathcal{G}_{\ell}}|\mathbf{W}[i, g]|$, $\overline{\boldsymbol{\beta}}_{\mathcal{I}}[i] = |{\boldsymbol{\beta}}_{\mathcal{I}}[i]|$ and $\overline{\boldsymbol{\beta}}_{\mathcal{G}}[\ell] = \max_{g \in \mathcal{G}_{\ell}} |{\boldsymbol{\beta}}_{\mathcal{G}}[g]|$. Only some brain regions are shown in this figure.\bigskip}
\label{matrixW}
\end{center}
\end{figure}

Regarding the matrix $\mathbf{W}$, the couples (brain region, gene) learnt through the task ``pMCI versus CN" are shown on Fig.  \ref{matrixW}. It can be seen that $\mathbf{W}$ has a sparse structure. Among the couples (brain region, gene) that have non null coefficients for the both tasks ``AD versus CN" and ``pMCI versus CN", there are (Left Hippocampus, MGMT), (Right Entorhinal, APOE) or (Left Middle Temporal, APOE). Only couples related to AD are selected by the model.

We noticed that genes and brain regions strongly related to AD are captured by the vectors ${\boldsymbol{\beta}}_{\mathcal{G}}$ and ${\boldsymbol{\beta}}_{\mathcal{I}}$, whereas genes less strongly related to AD are captured by the matrix $\mathbf{W}$. Coming back to original formulation described in section \ref{multdescr}, the contribution of the function $\alpha_0: \mathbf{x}_{\mathcal{G}} \mapsto {\boldsymbol{\beta}}_{\mathcal{G}}^{\top} \mathbf{x}_{\mathcal{G}}  + \beta_0$ is much smaller (in terms of weights) than the function $\boldsymbol{\alpha}: \mathbf{x}_{\mathcal{G}} \mapsto \mathbf{W} \mathbf{x}_{\mathcal{G}} + {\boldsymbol{\beta}}_{\mathcal{I}}$. Furthermore, Fig.  \ref{matrixW} shows that genetic data $\mathbf{x}_{\mathcal{G}}$ tend to express through $\mathbf{W}$, and thereby participate in the modulation of the vector ${\boldsymbol{\alpha}}(\mathbf{x}_{\mathcal{G}})$.


We compared our approach to \cite{wang_identifying_2012,aiolli_easymkl_2014}, for which the codes are available. The features that are selected by \cite{wang_identifying_2012,aiolli_easymkl_2014} are similar to ours for each modality taken separately. For instance, for \cite{wang_identifying_2012} and the task ``AD versus CN", SNPs that have the most important weights are in genes APOE (rs429358), BZW1 (rs3815501) and MGMT (rs7071424). However, the genetic parameter vector learnt from \cite{wang_identifying_2012} or \cite{aiolli_easymkl_2014} is not sparse, in contrary of ours. Furthermore, for \cite{aiolli_easymkl_2014}, the weight for the imaging kernel is nine times much larger than the weight for the genetic kernel.  These experiments show that the additive model with adapted penalties for each modality provides better performances than \cite{aiolli_easymkl_2014}, but our additive, multiplicative and multilevel models provide similar performances.

\section{Conclusion}

In this paper, we developed a novel approach to integrate genetic and brain imaging data for prediction of disease status. Our multilevel model takes into account potential interactions between genes and brain regions, but also the structure of the different types of data though the use of specific penalties within each modality. When applied to genetic and MRI data from the ADNI database, the model was able to highlight brain regions and genes that have been previously associated with AD, thereby demonstrating the potential of our approach for imaging genetics studies in brain diseases. 

\bigskip

\noindent \textbf{Acknowledgments.} The research leading to these results has received funding from the program \emph{Investissements d'avenir ANR-10-IAIHU-06}.
We wish to thank Theodoros Evgeniou for many useful insights. 

\begin{appendix}
\section{Probabilistic formulation}

This section proposes a probabilistic formulation for the model. The conditional probability is given by $p( y=1 | \mathbf{x}_{\mathcal{G}},   \mathbf{x}_{\mathcal{I}}) = \sigma\left(\mathbf{x}_{\mathcal{G}}^{\top}\mathbf{W}^{\top}\mathbf{x}_{\mathcal{I}} + \boldsymbol{\beta}_{\mathcal{I}}^{\top}\mathbf{x}_{\mathcal{I}} + \boldsymbol{\beta}_{\mathcal{G}}^{\top}\mathbf{x}_{\mathcal{G}} + \beta_0\right)$.
\begin{itemize}
\item For each region $i \in \mathcal{I}$ and gene $\mathcal{G}_{\ell}$, $\mathbf{W}_{i, \mathcal{G}_{\ell}} \sim \text{M-Laplace}(0, \lambda_W)$ (M-Laplace stands for ``Multi-Laplacian prior"). In other words:
\begin{align*}
p(\mathbf{W}; \lambda_W,   \mathcal{G},    \boldsymbol{\theta}_{\mathcal{G}}) & \propto \prod_{i=1}^{|\mathcal{I}|} \prod_{\ell=1}^{L} \text{e}^{-\lambda_W {\theta}_{\mathcal{G}_{\ell}} \| \mathbf{W}_{i,   \mathcal{G}_{\ell}} \|_2}
\end{align*}
\item For each region $i \in \mathcal{I}$,   $\boldsymbol{\beta}_i \sim \mathcal{N}\left(0, \frac{1}{2\lambda_{\mathcal{I}}} \right)$, i.e. $p(\boldsymbol{\beta}_{\mathcal{I}}; \lambda_{\mathcal{I}}) \propto \text{e}^{-\lambda_{\mathcal{I}} \|\boldsymbol{\beta}_{\mathcal{I}}\|_2^2}$
\item For each gene $\mathcal{G}_{\ell}$,   $\boldsymbol{\beta}_{\mathcal{G}_{\ell}} \sim \text{M-Laplace}(0,   \lambda_{\mathcal{G}})$, i.e. 
\[p(\boldsymbol{\beta}_{\mathcal{G}}; \lambda_{\mathcal{G}},   \mathcal{G},   \boldsymbol{\theta}_{\mathcal{G}}) \propto \prod_{\ell=1}^{L} \text{e}^{-\lambda_{\mathcal{G}} {\theta}_{\mathcal{G}_{\ell}} \| {\boldsymbol{\beta}}_{\mathcal{G}_{\ell}} \|_2}\]
\end{itemize}

Let $Y = (y^1, \ldots, y^N)$, $X_{\mathcal{I}} = (\mathbf{x}_{\mathcal{I}}^1, \ldots, \mathbf{x}_{\mathcal{I}}^N)$ and $X_{\mathcal{G}} = (\mathbf{x}_{\mathcal{G}}^1, \ldots, \mathbf{x}_{\mathcal{G}}^N)$. 

The generative model is given by:

\medskip

$p(\mathbf{W},   \boldsymbol{\beta}_{\mathcal{I}},   \boldsymbol{\beta}_{\mathcal{G}},  \beta_0,   Y,   X_{\mathcal{I}},   X_{\mathcal{G}}; \lambda_{W},  \lambda_{\mathcal{I}}, \lambda_{\mathcal{G}},   \mathcal{G},  \boldsymbol{\theta}_{\mathcal{G}})$
\vspace{-0.25cm}
\begin{align*}
\hspace{0.5cm} & \mathop{=}^{\text{Bayes}} 
p(Y,   X_{\mathcal{I}},   X_{\mathcal{G}}| \mathbf{W},   \boldsymbol{\beta}_{\mathcal{I}},   \boldsymbol{\beta}_{\mathcal{G}})  p(\mathbf{W}; \lambda_W,   \mathcal{G},   \boldsymbol{\theta}_{\mathcal{G}}) p( \boldsymbol{\beta}_{\mathcal{I}}; \lambda_{\mathcal{I}}) p(\boldsymbol{\beta}_{\mathcal{G}}; \lambda_{\mathcal{G}},   \mathcal{G},   \boldsymbol{\theta}_{\mathcal{G}})p(\beta_0)  \\
& \mathop{=}^{\text{obs iid}} 
\left( \prod_{k=1}^N p(y=y^k,   \mathbf{x}_{\mathcal{I}}^k,   \mathbf{x}_{\mathcal{G}}^k | \mathbf{W},    \boldsymbol{\beta}_{\mathcal{I}},   \boldsymbol{\beta}_{\mathcal{G}}) \right)   \\
&\hspace{1cm} p(\mathbf{W};\lambda_W,   \mathcal{G},   \boldsymbol{\theta}_{\mathcal{G}})p( \boldsymbol{\beta}_{\mathcal{I}}; \lambda_{\mathcal{I}}) p(\boldsymbol{\beta}_{\mathcal{G}}; \lambda_{\mathcal{G}},   \mathcal{G},   \boldsymbol{\theta}_{\mathcal{G}})p(\beta_0)  \\
& \propto \prod_{k=1}^N \sigma\left( {(\mathbf{x}_{\mathcal{G}}^{k})}^{\top} \mathbf{W}^{\top}\mathbf{x}_{\mathcal{I}}^k + \boldsymbol{\beta}_{\mathcal{I}}^{\top}\mathbf{x}_{\mathcal{I}}^k + \boldsymbol{\beta}_{\mathcal{G}}^{\top}\mathbf{x}_{\mathcal{G}}^k + \beta_0 \right)^{y^k} \\
& \quad \prod_{k=1}^N \left[1- \sigma\left( {(\mathbf{x}_{\mathcal{G}}^{k})}^{\top} \mathbf{W}^{\top}\mathbf{x}_{\mathcal{I}}^k + \boldsymbol{\beta}_{\mathcal{I}}^{\top}\mathbf{x}_{\mathcal{I}}^k + \boldsymbol{\beta}_{\mathcal{G}}^{\top}\mathbf{x}_{\mathcal{G}}^k + \beta_0 \right)\right]^{1-y^k} \\
&\quad \left( \prod_{i=1}^{|\mathcal{I}|} \prod_{\ell=1}^{L} \text{e}^{-\lambda_W  {\theta}_{\mathcal{G}_{\ell}} \| \mathbf{W}_{i,   \mathcal{G}_{\ell}} \|_2} \right) \times \text{e}^{-\lambda_{\mathcal{I}} \|\boldsymbol{\beta}_{\mathcal{I}}\|_2^2} \times \left( \prod_{\ell=1}^{L} \text{e}^{-\lambda_{\mathcal{G}} {\theta}_{\mathcal{G}_{\ell}} \| {\boldsymbol{\beta}}_{\mathcal{G}_{\ell}} \|_2}\right) 
\end{align*}

The \emph{maximum a posteriori} estimation is given by:
\begin{align*}
(\widehat{\mathbf{W}},   \widehat{\boldsymbol{\beta}}_{\mathcal{I}},   \widehat{\boldsymbol{\beta}}_{\mathcal{G}},   \widehat{\beta}_0) 
& \in \mathop{\text{argmax}}_{\mathbf{W},   \boldsymbol{\beta}_{\mathcal{I}},   \boldsymbol{\beta}_{\mathcal{G}},  \beta_0} p(\mathbf{W},   \boldsymbol{\beta}_{\mathcal{I}},   \boldsymbol{\beta}_{\mathcal{G}},  \beta_0 | Y,   X_{\mathcal{I}},   X_{\mathcal{G}}; \lambda_{W},  \lambda_{\mathcal{I}}, \lambda_{\mathcal{G}},   \mathcal{G},   \boldsymbol{\theta}_{\mathcal{G}})\\
& \in \mathop{\text{argmax}}_{\mathbf{W},   \boldsymbol{\beta}_{\mathcal{I}},   \boldsymbol{\beta}_{\mathcal{G}},  \beta_0} p(\mathbf{W},   \boldsymbol{\beta}_{\mathcal{I}},   \boldsymbol{\beta}_{\mathcal{G}},  \beta_0,   Y,   X_{\mathcal{I}},   X_{\mathcal{G}};\lambda_{W},  \lambda_{\mathcal{I}}, \lambda_{\mathcal{G}},   \mathcal{G},   \boldsymbol{\theta}_{\mathcal{G}})
\end{align*}

It is equivalent to minimize the function $S$ defined by:
\begin{align*}
S(\mathbf{W},   \boldsymbol{\beta}_{\mathcal{I}},   \boldsymbol{\beta}_{\mathcal{G}},   \beta_0) & = -\log p(Y,   \mathbf{W},   \boldsymbol{\beta}_{\mathcal{I}},   \boldsymbol{\beta}_{\mathcal{G}},  \beta_0 ,   X_{\mathcal{I}},   X_{\mathcal{G}}; \lambda_{W},  \lambda_{\mathcal{I}}, \lambda_{\mathcal{G}},   \mathcal{G},   \boldsymbol{\theta}_{\mathcal{G}}) \\
& = R_N(\mathbf{W},   \boldsymbol{\beta}_{\mathcal{I}},   \boldsymbol{\beta}_{\mathcal{G}},   \beta_0) + \Omega(\mathbf{W},   \boldsymbol{\beta}_{\mathcal{I}},   \boldsymbol{\beta}_{\mathcal{G}}) 
\end{align*}

 
\end{appendix}

\bibliographystyle{plain} 

\end{document}